# Monocular Vehicle Self-localization method based on Compact Semantic Map

Zhongyang Xiao, Kun Jiang, Shichao Xie, Tuopu Wen, Chunlei Yu, Diange Yang

*Abstract*—High precision localization is a crucial requirement for the autonomous driving system. Traditional positioning methods have some limitations in providing stable and accurate vehicle poses, especially in an urban environment. Herein, we propose a novel self-localizing method using a monocular camera and a 3D compact semantic map. Pre-collected information of the road landmarks is stored in a self-defined map with a minimal amount of data. We recognize landmarks using a deep neural network, followed with a geometric feature extraction process which promotes the measurement accuracy. The vehicle location and posture are estimated by minimizing a self-defined re-projection residual error to evaluate the map-to-image registration, together with a robust association method. We validate the effectiveness of our approach by applying this method to localize a vehicle in an open dataset, achieving the RMS accuracy of 0.345 meter with reduced sensor setup and map storage compared to the state of art approaches. We also evaluate some key steps and discuss the contribution of the subsystems.

## I. INTRODUCTION

Vehicle self-localization is a fundamental component in realizing intelligent vehicle's planning, controlling, V2X, and other location-based applications. In recent decades, researchers have been making efforts to improve localization accuracy and stability, but existing technology is still unable to meet the needs of autonomous driving, especially on urban roads[1]. In this paper, we propose a high-precision self-localization solution in the context of urban driving.

Amongst the existing positioning methods, Differential Global Position System (DGPS) with Real-time Kinematic data (RTK) can achieve centimeter-level precision in open, wild ground[2]. But in urban environments, signal reflection and blockage by building, trees, and traffic facilities can easily cause high positioning error. Dead Reckoning (DR) is widely used as a supplement to GNSS, but in the long-period inavailability of GNSS signal, DR may have large location drift due to error accumulation[3].

Map-based localization can eliminate the location drift, in which information obtained by onboard sensors is registered to the pre-collected map to give a relative position on the map. In the high precision localization, most works use dense and detailed maps collected by laser scanner or camera [4][5]. These maps contain rich features, which is beneficial for the alignment with on-board sensors. However, they are far from mass application because of the large map data size and the poorly organized map model which is hard to update. As for sensors mounted on vehicles, many works adopt Lidar[6], radar[7], or stereo vision[8] that can give a range measurement between the sensor and landmarks on the road. However, the monocular camera cannot acquire depth information directly, thus its application is much limited in the map-based localization.

In this paper, we propose a self-localizing approach by matching monocular vision with a lightweight 3D map. In particular, we define a compact semantic map model to describe road landmarks with high precision. We extract semantic and geometric features from the camera with a deep learning method, and align the features with the map data taking mismatching into account as well. We locate the ego-vehicle by minimizing a residual registration error, and validate our method by showing the experimental results on KITTI [9], a publicly available urban benchmark dataset.

The main contributions of our work are as follows:
(A) A self-defined compact map model which requires less storage space. A corresponding landmark preselection method is also proposed to simplify the alignment with onboard measurement.
(B) A monocular localization method based on optimizing a multi-model re-projection error. We adopt heuristic measures to improve the convergence dealing with the sparse landmarks case.
(C) A simultaneous association and localization method, which is robust to error or leak detection in the image processing.

## II. RELATED WORKS

In the past few years, many researchers treat the Map-based Localization problem as a retrieval or association problem between pre-built "map" and onboard sensors. Caselitz et al. [4] track the camera pose by matching the reconstructed 3D point cloud with a 3D Lidar map, achieving a translational error of $0.3 \pm 0.11$ meter. Ryan et al. [10] localize the camera by maximizing the Mutual Information (MI) between the captured images with a re-projected picture from 3D map. Stenborg et al. [11] align measurement of two cameras with a semantically segmented 3D point cloud. The maps they used are raw or down-sampled 3D Lidar point clouds, which is poorly organized and very large in data size. Sefati et al. [8] align a compact 2D roadmap to a stereo vision system, with the accuracy of 50 centimeters on flat roads. Schindler [12] propose a localizing method by matching lane markers detected by camera and laser scanner with a lane level map. These works attempt to capture the relative position of landmarks with distance measuring sensors. Different from all these works, the proposed method uses only one camera with a lightweight 3D semantic map.

The most traditional localization method is the Monte Carlo Location (MCL), which estimates the vehicle's pose and position by calculating the posterior probability. The prior probability is based on the prediction model and the likelihood is formulated by the sensor measurement. Particle Filter (PF) is applied to recursively update the estimation, in which the probability distribution is simulated by particles. MCL is typically applied in scenarios where the vehicle moves in a planner as the DOF is limited to 3 (i.e. the longitude, lateral position and the orientation). Unfortunately, MCL does not apply to high DOF, which will cause the sharp increase of particles number. Instead of MCL, we address the problem in an optimization method, with the camera's pose described in 6 DOF, which describes the full motion of the vehicle.

"Landmarks" (or features in some research) are the clue to the association of maps and sensor data. In most works of camera-based localization, graphics features like SIFT, SURF ORB points or Edge features are extracted from the image. The extraction of these features is relatively fast, but they are sensitive to the change of view, light condition and occlusion. Other works detect lane markings, trees, or lamp poles with role-based method[8][12]. It is well known that the generalization ability of these rule-based methods is lower. In this work, we extract geometrical features with semantic based on deep learning method, which is proven more accurate and robust[13]. Admittedly, the deep learning method has a higher calculation amount than the traditional ways, but with the rapid development of deep learning and vehicular computing hardware, the method we used can run in real time[14].

The problem of Map-based Localization of ego vehicle is similar to the re-localization or loop closure detection of simultaneous localization and mapping (SLAM) problem in the robotic domain. We are inspired a lot by the theories of SLAM, e.g. the description of camera pose, the optimization method of solving the Bundle Adjustment (BA) problem and the semantic SLAM which has exploded in recent years[15]. However, the problem we meet in this paper is different from typical SLAM re-localizing problem. When a robot revisits a place, it can locate itself in the pre-built map by detecting loop closure[16]. In existing SLAM methods, types of the pre-built map and onboard sensor should be similar, and the interval between the two visits should not be long, as any change of environment is unfavorable for the re-localizing. In contrast, the map data in our method is obtained by Lidar while the onboard sensor is a monocular camera, and the map is obtained long before the re-localizing is executed. Also, the features we used as constraint of BA problem is much sparser than those used in typical SLAM works, which requires us to design a matching algorithm applicable to traffic scenes.

III. OVERVIEW OF THE APPROACH

In our approach, we localize the vehicle by estimating the pose of the onboard camera in the map coordinate, i.e., the earth coordinate. Therefore, the localization task comes down to a 3D-2D (map to image) registration problem. An overview of our proposed method is shown in Fig. 1.

We form a 3D map by splicing multi-frame Lidar point cloud and extracting lane lines, pole-like objects (e.g. lamp poles, milestones), traffic signs, using the Point Cloud Library (PCL)[17]. According to a self-defined map model, the semantic and geometric information of the landmarks is organized and stored. Landmarks that are likely to be matched with onboard measurement are extracted according to a map preselection approach.

Semantic segmentation is conducted on the captured image with the PSPNet[13] and a self-defined neural network, where pixels are classified according to the semantic. Then we extracted lines and points features with region growing and line fitting with RANSAC method.

The preselected landmarks from 3D map are projected according to a supposed camera pose, and a cost function is calculated considering the re-projection error of these landmarks and their aligned features on the captured image. We estimate the camera pose iteratively by minimizing the cost function with Levenberg–Marquardt algorithm. To improve the convergence of the optimization, we also added a term in the cost function. The initial value for optimization is calculated based on a rough location and the 3D map.

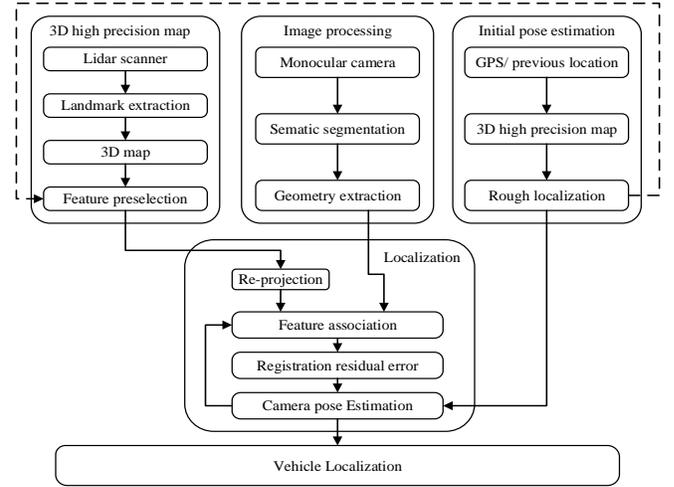

Figure 1.   Overview of the proposed method

IV. METHOD

A. Semantic map and landmark preselection

In the existing high precision map standard, for example, the Navigation Data Standard (NDS), only positions of bounding boxes of localizing objects are stored, which cannot provide landmark information with enough precision. In our research, we define a new map model organizing both semantic and precise geometrical information.

The 3D map is expressed as,

$$\mathbf{M}_a = (\mathbf{L}_a, \mathbf{P}_a) \quad (1)$$

where, $\mathbf{L}_a$ is the set of line landmarks like lane lines, pole-like objects, and $\mathbf{P}_a$ is the set of point landmarks like traffic signs.

The component of line landmarks is defined as,

$$\mathbf{l}_a = (\mathbf{l}_m, s_l, m_l, r_l) \quad (2)$$

where, $\mathbf{l}_m = (\mathbf{p}_{l1}^T, \mathbf{p}_{l2}^T)^T \in \mathbb{R}^6$, $\mathbf{p}_{l1}$ and $\mathbf{p}_{l2}$ are two control points on the landmark in the map coordinate system, which are extracted by fitting the point cloud of the objects' straight parts. Take lamp poles for example, as is shown in Fig. 2 (a), the points are from the Lidar scanner. We fit points of the straight part with a line $l$ in 3D. We make two planes perpendicular to $l$ over the most vertical points $p_1$ and $p_2$ along the straight line, and the intersection points $\mathbf{p}_{l1}$ and $\mathbf{p}_{l2}$ are determined as the control points.

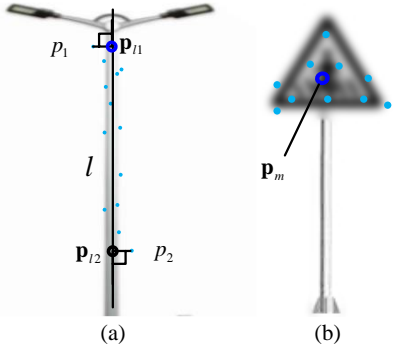

Figure 2. Control points definition

$s_l$ is the object's semantic and $m_l$ is the rough size of the object, which is the length of the line segment of the object. $r_l$ is the index of road on which the object is attached, which is stored for landmark preselection.

The component of point landmarks is defined as,

$$\mathbf{p}_a = (\mathbf{p}_m, s_p, m_p, r_p) \quad (3)$$

where, $\mathbf{p}_m$ is the centroid of landmark point cloud as is shown in Fig.2 (b), and $s_p$ is the semantic. $m_p$ is determined as the largest distance between two points in the point cloud, and $r_p$ is the road index of the point landmark.

When using the 3D map, not all features are extracted for alignment – only those are likely to be captured by the camera are used. We filter landmarks according to the road index, and compute $s$ for each extracted landmark:

$$s = \frac{m}{\|\bar{\mathbf{p}} - \mathbf{p}\|_2} \quad (4)$$

where, $\bar{\mathbf{p}}$ is the rough coordinate, $\mathbf{p}$ is the control point, and $m$ is the rough size. If $s$ exceeds a certain threshold (we take 0.017 in this work), the landmark is extracted from the map to the preselected set.

The case of extracting lane lines is a little different. We store each lane line in the map with a point sequence and extract the points 5~20 meters in front of the rough coordinate $\bar{\mathbf{p}}$. We use a straight line to fit these points and form the line landmark with the same method of extracting pole-like landmarks.

### B. Semantic geometric features extraction

We imply two steps for processing the captured image: semantic segmentation and geometric feature extraction. We define a network to classify lanes as shown in Fig. 3. This network uses the structure similar to vgg16 to subsample the image, transforms with the bottleneck of RESnet in the middle, and uses the bilinear interpolation to upsample the image. To classify road lamps, traffic signs and so on, we use a trained PSPNet with cityscapes dataset [13]. Finally, we get several predicted feature maps for each semantic class.

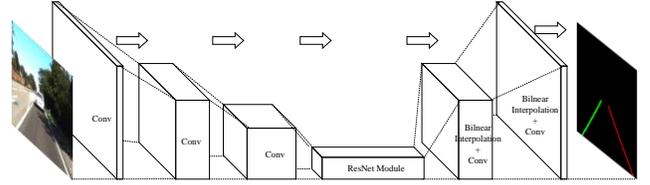

Figure 3. Network for lane lines detection

For a specific semantic class, we transform the corresponding final feature map into a two-value image. As shown in figure 4(b), take pole-like objects for example, pixels on the feature map of which the probability exceed a certain threshold (0.1 in our test) are transformed to 1, and the left pixels are transformed to 0. A region-growing method is applied to divide the two value image into different regions as is shown in figure 3(c). Each region is fitted to a line using RANSAC method. The geometric extraction result of lamp poles is shown with red line in figure 3(d). As for and traffic signs, the control point is calculated as the center of the region.

The set of extracted lines is denoted as $\mathbf{L}_r = \{\mathbf{l}_1...\mathbf{l}_{N_l}\}$, where $\mathbf{l}_i = (\mathbf{P}_{i1}^T, \mathbf{P}_{i2}^T)^T \in \mathbb{R}^4$, and $\mathbf{P}_{i1}$, $\mathbf{P}_{i2}$ are the control points of the $i$-th line feature recognized on the image. Similarly, the set of extracted points is denoted as $\mathbf{P}_r = \{\mathbf{p}_1...\mathbf{p}_{N_r}\}$, where $\mathbf{p}_j \in \mathbb{R}^2$ is the coordinate of the $j$-th point feature on the current image.

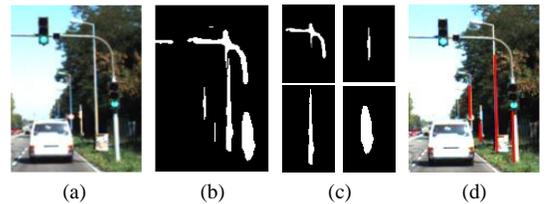

Figure 4. Semantic geometry extraction

### C. The localization problem and camera model

As the camera is mounted on the vehicle, the problem of vehicle self-localization is deduced to the estimation of the camera pose relative to the map. As is shown in Fig. 5, we use a 6-DOF camera pose model $\mathbf{p}_c = (C_x, C_y, C_z, \phi, \theta, \psi)^T$ to describe the camera pose, where $C_x, C_y, C_z$ are the three coordinates of the camera center in the map coordinate system, and $\phi, \theta, \psi$ are the yaw, pitch, and roll angle of the camera coordinate system relative to the map coordinate system.

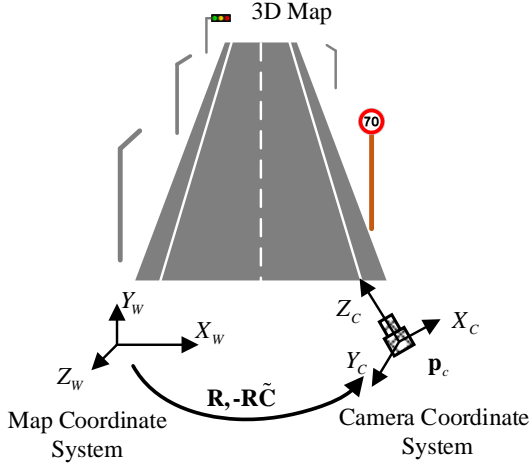

Figure 5. Camera pose estimation

We imply a finite projective camera model[18] to define the camera projection function $\Pi : (\mathbb{R}^3, \mathbb{R}^6) \mapsto \mathbb{R}^2$ that projects the 3D homogeneous point $\mathbf{X} = (X, Y, Z, 1)^T$ in the map coordinate to the 2D image coordinate $\mathbf{u} = (u, v, 1)^T$ known the 6-DOF camera pose $\mathbf{p}_c$:

$$\Pi : \mathbf{u} = \mathbf{KR}[\mathbf{I} \mid -\tilde{\mathbf{C}}]\mathbf{X} \quad (5)$$

In (5), $\mathbf{K}$ is the internal parameters of the camera, $\tilde{\mathbf{C}}$ is the coordinate of the camera center in the map coordinate system, and $\mathbf{R}$ is the relative rotation matrix from the map coordinate system to the camera coordinate system. Based on $\Pi$, we define a point projection function $\pi_P : (\mathbb{R}^3, \mathbb{R}^6) \mapsto \mathbb{R}^2$ and a line projection function $\pi_L : (\mathbb{R}^6, \mathbb{R}^6) \mapsto \mathbb{R}^4$ projects the 3D point and line features in the map to the image domain given the camera pose $\mathbf{p}_c$:

$$\mathbf{u} = \pi_P(\mathbf{p}_m, \mathbf{p}_c) \quad (6)$$

$$\mathbf{l} = \pi_L(\mathbf{l}_m, \mathbf{p}_c) \quad (7)$$

where $\mathbf{p}_m \in \mathbb{R}^3$ denotes the center of a point landmark, and $\mathbf{l}_m \in \mathbb{R}^6$ denotes the line landmark as described in section IV(A). In (7), the two control points in $\mathbf{l}_m$ is respectively projected to the image coordinate according to $\Pi$. We use $\mathbf{l} = (\mathbf{u}_1^T, \mathbf{u}_2^T)^T \in \mathbb{R}^4$ to describe the projected line landmark in the image coordinate system.

### D. Lines and points alignment based pose estimation

The basic idea of estimating the camera pose is to maximize the similarity between the projected features and the current image frame. We propose a residual model based on both lines and points features' distance, given the extracted map $\mathbf{m}_e$, the recognized lines and points features $\mathbf{L}_r$, $\mathbf{P}_r$ with their association $\mathbf{c}$:

$$r(\mathbf{m}_e, \mathbf{L}_r, \mathbf{P}_r, \mathbf{c}, \mathbf{p}_c) = \sum_{c_l^i \in \mathbf{c}_l} D_L^2(\pi_L(\mathbf{L}_m^i, \mathbf{p}_c), \mathbf{l}_{c_l^i})$$
$$+ \sum_{c_P^j \in \mathbf{c}_P} D_P^2(\pi_P(\mathbf{P}_m^i, \mathbf{p}_c), \mathbf{p}_{c_l^i}) \quad (8)$$

Here we define the correspondence set $\mathbf{c} = (\mathbf{c}_l, \mathbf{c}_p)$ of features on the map and the current image, where the line correspondence $\mathbf{c}_l = \{c_l^1, c_l^2, ...\}$, $c_l^i \in \{1, ..., N\}$ ($N$ is the number of line features on the current image), and $c_l^i = j \leq N$ if the $i$-th line feature in the pre-selected landmarks correspond to the $j$-th line feature on the current image. The point correspondence set $\mathbf{c}_p$ is defined in the same way. In this section, we assume that the lines and points feature correspondence is given, and the association problem will be addressed in the next part.

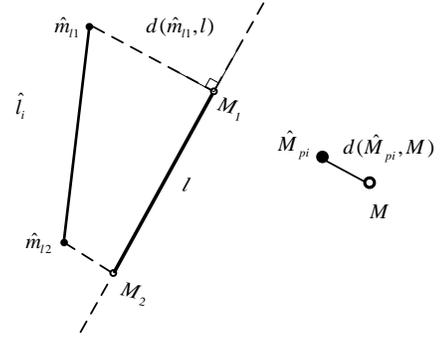

Figure 6. Definition of distance between lines and points

We explain the definition of the distance between two lines $D_L : (\mathbb{R}^4, \mathbb{R}^4) \mapsto \mathbb{R}$ in the image coordinate system in Fig. 6. $\hat{l}_i \in \mathbb{R}^4$ is a projected line from a 3D feature in the map following (5), and $l \in \mathbb{R}^4$ is an extracted line in the current image with two endpoints $M_1$ and $M_2$. The distance of $\hat{l}_i$ and $l$ is defined by the following equation:

$$D_L(\hat{l}_i, l) = \frac{1}{2} \sum_{j=1}^{2} \frac{\left| [M_1 - M_2 \mid \hat{m}_{li} - M_1] \right|}{\|M_1 - M_2\|_2} \quad (9)$$

where, $\hat{m}_{l1}$ and $\hat{m}_{l2}$ are the projected point of the two 3D line feature endpoints. And the distance of two aligned points is calculated with $D_P : (\mathbb{R}^2, \mathbb{R}^2) \mapsto \mathbb{R}$:

$$D_P(\hat{M}_{pi}, M) = \|\hat{M}_{pj} - M\|_2 \quad (10)$$

where $\hat{M}_{pj}$ is the projected point landmark and $M$ is its corresponding point recognized in the current frame.

In the urban driving scene, we assume that the vehicle seldom leaves the ground, and the ground is flat in most cases. To improve the convergence of the optimization, especially in sparse landmark case, we add a "soft constraint" to pitch, roll angle of the camera $\theta, \psi$ and the height of the camera $C_y$, by adding one item to the cost function:

$$r_n(\mathbf{p}_c) = \|(\theta, \psi, C_y - (y_{lane} + H))\|_2^2 \quad (11)$$

where, $y_{lane}$ is the height of the nearest control point of lanes extracted from the map, H is the height of the camera off the ground. We denote the whole residual as:

$$R(\mathbf{p}_c) = \sum_{c_l^i \in \mathbf{c}_l} D_L^2(\pi_L(\mathbf{L}_m^i, \mathbf{p}_c), \mathbf{l}_{c_l^i}) + \sum_{c_p^j \in \mathbf{c}_p} D_P^2(\pi_P(\mathbf{P}_m^i, \mathbf{p}_c), \mathbf{p}_{c_l^j})$$
$$+ \lambda_n^2 \|(\theta, \psi, C_y - (y_{lane} + H))\|_2^2 \quad (12)$$

where $\lambda_n$ is the weight of the added item in the overall cost function, which is set to 0.001 in our test (the length is in centimeters and the angle is in angular degree). The optimal estimation of the camera pose relative to the map can be obtained by solving the optimization problem (13). In our work we address the problem with Levenberg-Marquart method.

$$\mathbf{p}_c^* = \underset{\mathbf{p}_c}{argmin}[r(\mathbf{m}_e, \mathbf{L}_r, \mathbf{P}_r, \mathbf{c}, \mathbf{p}_c) + \lambda_n^2 r_n(\mathbf{p}_c)] \quad (13)$$

*E. Feature association*

The association of features of the map with the current image is determined simultaneously with localization. We establish the hypothesis association considering both the semantic and the re-projecting distance, and validate the association by analyzing the result of localization that uses the association.

As is shown in Algorithm *Closest_correspond*, we determine the association given the camera pose, map, image recognition result and the threshold of the distance. The extracted map is projected according to the camera pose, and we assume that the map landmarks' nearest feature in the current image is their associated feature if the distance is smaller than the threshold.

The progress of association and localization is shown in Algorithm *Associate_and_localize*. We find an initial correspondence set $\mathbf{c}^0$ according to the initial pose $\mathbf{p}_c^0$. From $\mathbf{c}^0$, four line and one point correspondences are *randomly* selected each time as "hypothesis association". We validate the association by estimating the camera pose $\hat{\mathbf{p}}_c^*$ according to the correspondence set. The validation rule is based on our observation that if the hypothesis association is right matching, the residual function is relatively small and the estimated pose is not too far from the initial guess. Once a good pose $\hat{\mathbf{p}}_c^*$ is found, we establish a new correspondence set $\breve{\mathbf{c}}$ based on $\hat{\mathbf{p}}_c^*$. If the residual and the estimated pose meet a certain condition as is shown in line 6 and 9, the new correspondence set is assumed as a "perfect association", and the final localization result is given by optimizing based on the "perfect association". The reason why we do another optimization is that compared to the $\hat{\mathbf{p}}_c^*$, the final localization result is calculated based on more good correspondences, which makes it more precise.

1. **Algorithm** *Closest_correspond*$(\mathbf{m}_e, \mathbf{L}_r, \mathbf{P}_r, \mathbf{p}_c, D_1, D_2)$:
2.   *for* $\mathbf{l}_m^i$ in $\mathbf{m}_e$
3.     $\hat{\mathbf{l}}^i = \pi_L(\mathbf{l}_m^i, \mathbf{p}_c)$
4.     In $\mathbf{L}_r$, find the closest line of $\hat{\mathbf{l}}^i$ in the same semantic: $\mathbf{l}_r^j$
5.     *if* $D_L(\hat{\mathbf{l}}^i, \mathbf{l}_r^j) \leq D_1$
6.       $c_l^{i0} = j$ , $\mathbf{c}_l^0 = \mathbf{c}_l^0 \cup c_l^{i0}$
7.     *endif*
8.   *endfor*
9.   *for* $\mathbf{p}_m^i$ in $\mathbf{m}_e$
10.     $\hat{\mathbf{p}}^i = \pi_P(\mathbf{p}_m^i, \mathbf{p}_c)$
10.     In $\mathbf{P}_r$, find closest point of $\hat{\mathbf{p}}^i$ with the same semantic: $\mathbf{p}_r^j$
11.     *if* $D_L(\hat{\mathbf{p}}^i, \mathbf{p}_r^j) \leq D_2$
12.       $c_p^{i0} = j$ , $\mathbf{c}_p^0 = \mathbf{c}_p^0 \cup c_p^{i0}$
13.     *endif*
14. *endfor*

1. **Algorithm** *Associate_and_localize*$(\mathbf{m}_e, \mathbf{L}_r, \mathbf{P}_r, \mathbf{p}_{c0})$
2.   $\mathbf{c}^0 = $ *Closest_correspond*$(\mathbf{m}_e, \mathbf{L}_r, \mathbf{P}_r, \mathbf{p}_c^0, D_1, D_2)$
3.   *for* $c_l^{1-4}$ and $c_p^1$ in $\mathbf{c}^0$
4.     $\hat{\mathbf{c}} = (c_l^1, c_l^2, c_l^3, c_l^4, c_p^1)$
5.     $\hat{\mathbf{p}}_c^* = \underset{\mathbf{p}_c}{argmin}[r(\mathbf{m}_e, \mathbf{L}_r, \mathbf{P}_r, \hat{\mathbf{c}}, \mathbf{p}_c) + \lambda_n^2 f_n(\mathbf{p}_c)]$
6.     *if* $\sqrt{R(\hat{\mathbf{p}}_c^*)} \leq R_1 \&\& \|\hat{\mathbf{p}}_c^* - \mathbf{p}_c^0\|_2 \leq D_5$
7.       $\breve{\mathbf{c}} = $ *Closest_correspond*$(\mathbf{m}_e, \mathbf{L}_r, \mathbf{P}_r, \mathbf{p}_c^0, D_3, D_4)$
8.       $\mathbf{p}_c^* = \underset{\mathbf{p}_c}{argmin}[r(\mathbf{m}_e, \mathbf{L}_r, \mathbf{P}_r, \breve{\mathbf{c}}, \mathbf{p}_c) + \lambda_n^2 f_n(\mathbf{p}_c)]$
9.       *if* $\sqrt{R(\mathbf{p}_c^*)} \leq R_2 * size(\hat{\mathbf{p}}_c^*) \&\& size(\breve{\mathbf{c}}) \geq 0.5 size(\mathbf{c}^0)$
10.         return $\mathbf{p}_c^*$, $\breve{\mathbf{c}}$
11.         break
12.       *endif*
13.     *endif*
14. *endfor*

In our test, the initial pose of camera is given by the GPS in the first two frames and determined by a unique speed assumption in the following frames: $\mathbf{p}_c^0 = 2^{k-1}\mathbf{p}_c^* - {}^{k-2}\mathbf{p}_c^*$, where ${}^{k-1}\mathbf{p}_c^*$ and ${}^{k-2}\mathbf{p}_c^*$ are the localization result of the last two frames. As for the thresholds in the method, we set $D_1 = D_2 = 300$ and $D_3 = D_4 = 10$, $D_5 = 30$, $R_1 = 200$, $R_2 = 4$.

V. EXPERIMENTAL RESULTS AND DISCUSSION

To evaluate the proposed method, we conduct an experiment on the 4th sequence of KITTI odometry dataset, which contains Lidar data frames and their sync image frames. We generate a map with the point clouds of Lidar scanner, and extract lines and points features according to our visual processing method. The localization method is evaluated with 190 frame images during 270 meters travel. In this section, we will present and discuss the result of some critical intermediate steps and the final evaluation of the localization method.

*A. Map creation*

We take the coordinate system of the first frame left camera as the map coordinate system. In producing the map, we splice point clouds generated by an HDL-64E Lidar in the dataset with 5 frames as intervals after transforming Lidar frames to the first map coordinate according to the pose provided in the dataset. We manually select regions of interest (ROI) of lamp poles, milestones, lane lines, and traffic signs and extract their point cloud with PCL tools. The raw data of the map and the ROI results are shown in Fig.7 (a), while the extracted semantic map is shown in Figure 7(b). It is worth mentioning that, the Lidar only sweep the lower part of the lamp, so the lamp poles in our map look shorter than the real ones. We extract 21 pole-like objects, 5 traffic signs and 2 lane lines in our test. The cost of storage of our map in ASCII format is 4KB, which is much less than the cost of a raw or down-sampled 3D Lidar cloud map of the same road (usually over MB level).

## C. Feature association result

We give the number of correspondences we estimate in every frame, which are used to finally estimate the camera pose. In each frame, we find 4 to 11 lines and points correspondences, with average number of 5.7. The influence of correspondence amount on positioning accuracy is discussed in the next section.

Examples of feature association result are shown in Fig. 10 intuitively with white boxes while the image features are in blue and re-projected map features (according to final localization results) are in red. There are some error and leak detection in the image processing step, but our method successfully avoids matching them with the map landmarks.

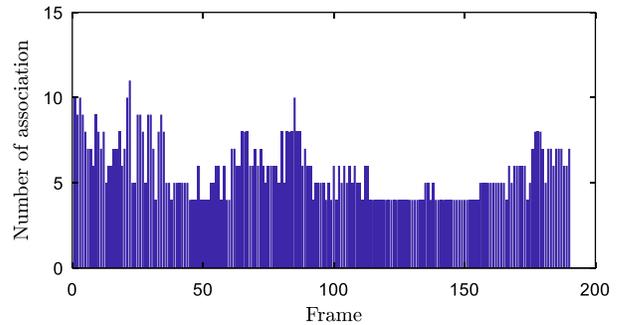

Figure 9. Number of association estimated in every frame

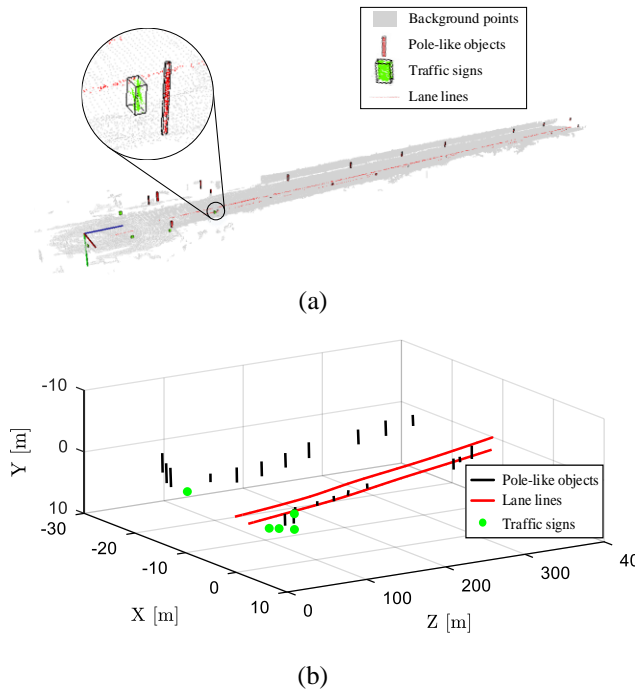

(a)

(b)

Figure 7. Map generation result

## B. Image processing

We show examples of images and segmentation output of PSPnet and our deep network in Fig. 8, while the feature extraction result is shown with red lines in Fig. 10. We can see that the result of feature extraction is more accurate than that of segmentation, which is advantageous to our localization.

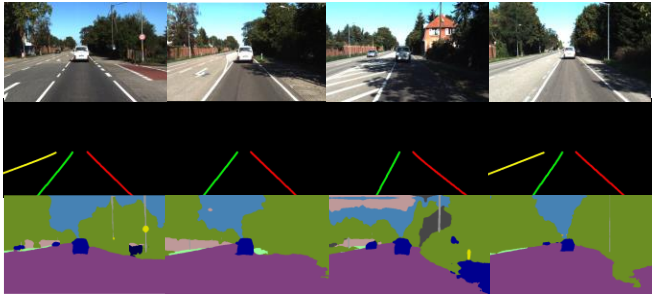

Figure 8. Example images and segmentation results

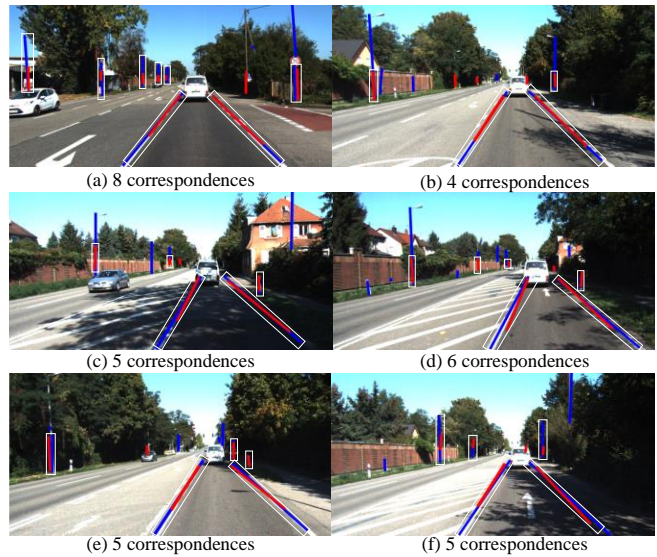

(a) 8 correspondences    (b) 4 correspondences
(c) 5 correspondences    (d) 6 correspondences
(e) 5 correspondences    (f) 5 correspondences

Figure 10. Feature association result. White boxes: correspondences estimated; Blue lines: recognized features; Red lines: Projected features from the vector map.

## D. Localization result

The localizing results and the ground truth given in the dataset are shown in Fig. 11, while the localizing error is shown in Fig. 12. The RMS positioning error is 0.345 meter, and 90.00% of the positioning errors are below 0.5 meter. The average of angle error is 0.019 rad, with the maximum of 0.11 rad. We also compare the RMS localization error of two state of art map-based methods in Table I. In general, our method gains a relatively high accuracy using low-cost sensors and lightweight map.

In our test, we find two frames (frame 92 and 99) with position error over 1 meter. Three possible facts may lead to the large error: 1. The landmarks are extremely sparse (only two lamp poles and two lane lines are available for association). 2. The lane recognition result is inaccurate when crossing the shade of trees. 3. The pose of Lidar which generates the map may be not accurate, as the location of Lidar in the dataset is obtained fusing a GPS. However, this problem is rarely seen. There are other places with sparse landmarks, but the localization error is small.

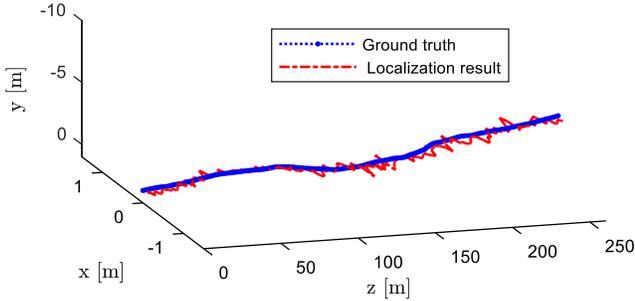

Figure 11. Localizing results and the ground truth

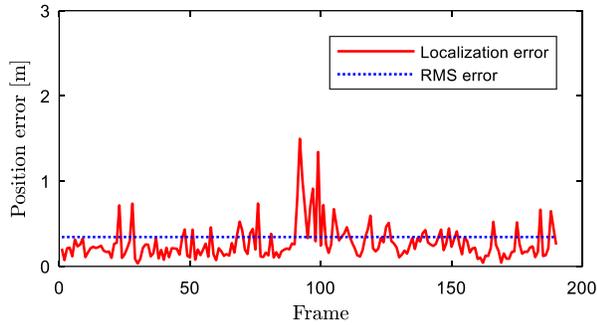

Figure 12. Positioning accuracy of our method

TABLE I. COMPARISON OF RMS LOCATION ERROR

| Method | Caselitz's | Andreas Schindler's | Erik Stenborg's | Ours |
|---|---|---|---|---|
| Sensors onboard | 1 camera | 1 camera +laser canner + IMU | 2 cameras | 1 camera & vector map |
| Map used | 3D lidar map at resolution of 20cm | Vector map | 3D segmented cloud map | Compact semantic map |
| RMS location error | ~0.3m[4] | <1m[12] | ~0.6m[11] | 0.345m |

### E. Discussion on the optimizing process

To illustrate the process of the map-to-image registration clearly, we give the shots of one optimization in Fig. 13. This optimization is based on 6 correspondences, and the whole process take 4 iterating times. In this figure, the localization error and norm of residual vector in Levenberg-Marquart process are shown. We can see that features in the image is gradually aligned to the landmarks projected to 2D.

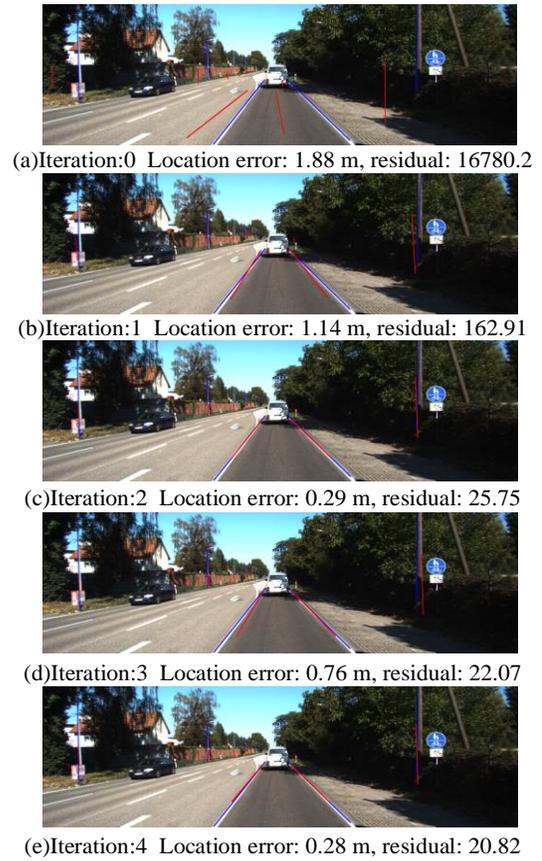

(a)Iteration:0  Location error: 1.88 m, residual: 16780.2

(b)Iteration:1  Location error: 1.14 m, residual: 162.91

(c)Iteration:2  Location error: 0.29 m, residual: 25.75

(d)Iteration:3  Location error: 0.76 m, residual: 22.07

(e)Iteration:4  Location error: 0.28 m, residual: 20.82

Figure 13. Process of map matching. Blue lines: recognized features; Red lines: Projected features from the map.

We show three projecting components of the localization error in Fig. 8. Apparently, the lateral and vertical accuracy is higher than the longitudinal one. To explain this, we show the "terrain" of the cost function at the vicinity of truth value as is shown in Fig. 15 (a) and (b) (when showing the cost function of two dimension, parameters of the other dimensions are set to true values). In Fig. 15 (a), we can see that the surface of cost function is steeper in lateral direction than that in longitudinal one. The reason why the vertical error is smaller is similar. We also show the sensitivity of the cost function to the angle error in Fig. 15 (c) and (d). It can be seen that it is more sensitive to yaw angle, which is just changing frequently during driving.

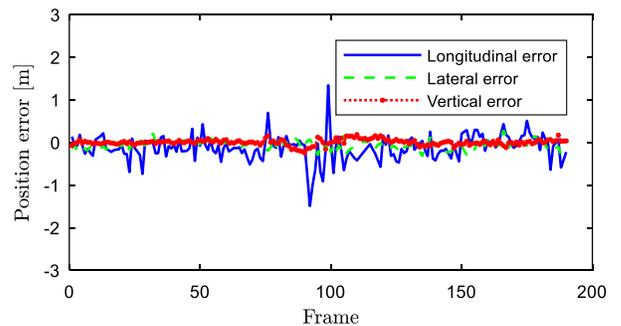

Figure 14. Lateral, longitudinal and vertical accuracy

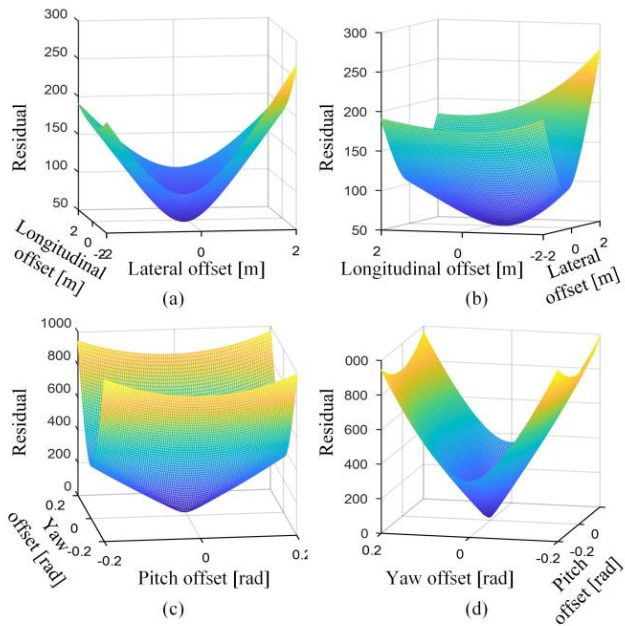

Figure 15.  Analyse of convergence in different directions

To sum up, our method achieves the localizing accuracy of state of art, and compared to other work of map-based localization, the onboard map of our work is much more lightweight. Except the first two frames, we only use one monocular camera as the onboard sensor. In the whole following positioning, our method does not depend on satellite communication, which makes it more suitable for urban road.

## VI. CONCLUSION

In this work, we propose a vehicle self-localization method based on a monocular camera and a self-defined compact high precision roadmap. The map is built to store prior information about road landmarks, with consideration of storage space and landmark pre-selection. In image processing, we extract semantic geometry features using deep learning followed by a line or point fitting method. The vehicle pose is estimated by optimizing a self-defined re-projecting residual error, simultaneously establishing the association of landmarks in the map with features of measurement.

We give the evaluation of our method in the KITTI odometry dataset with a RMS error of 0.345 meter. Evaluation of intermediate subsystems in our method is also presented. Comparing to other state of art works, our method has a much lower cost and map storage, which shows a promising prospect in localizing a vehicle independently. It could also be considered as an incremental method to existing approaches.

In further works, we will denoise the result with filter and consider data fusion with other localization methods. We will also improve the model of map and image processing to handle the case of driving on curves.

## REFERENCES

[1] Brubaker, Marcus A., A. Geiger, and R. Urtasun. Map-Based Probabilistic Visual Self-Localization. IEEE Computer Society, 2016.
[2] Montemerlo, Michael, et al. "Junior: The Stanford entry in the Urban Challenge." *Journal of Field Robotics* 25.9(2008):569–597.
[3] Ojeda, Lauro, and J. Borenstein. "Personal Dead-reckoning System for GPS-denied Environments." *IEEE International Workshop on Safety, Security and Rescue Robotics* IEEE, 2007:1--6.
[4] Caselitz, Tim, et al. "Monocular camera localization in 3D LiDAR maps." *Ieee/rsj International Conference on Intelligent Robots and Systems* IEEE, 2016:1926-1931.
[5] Burgard, W, O. Brock, and C. Stachniss. "Map-Based Precision Vehicle Localization in Urban Environments." *Robotics: Science & Systems Iii, June, Georgia Institute of Technology, Atlanta, Georgia, Usa* DBLP, 2007:121-128.
[6] Levinson, J, and S. Thrun. "Robust vehicle localization in urban environments using probabilistic maps." *IEEE International Conference on Robotics and Automation* IEEE, 2010:4372-4378.
[7] Ward, Erik, and J. Folkesson. "Vehicle localization with low cost radar sensors." *Intelligent Vehicles Symposium* IEEE, 2016:864-870.
[8] Sefati, M., et al. "Improving vehicle localization using semantic and pole-like landmarks." *Intelligent Vehicles Symposium* IEEE, 2017.
[9] Geiger, Andreas. "Are we ready for autonomous driving? The KITTI vision benchmark suite." *IEEE Conference on Computer Vision and Pattern Recognition* IEEE Computer Society, 2012:3354-3361.
[10] Eustice, Ryan M, and R. W. Wolcott. "Visual Localization Within LIDAR Maps.", US20160209846. 2016.
[11] Stenborg E, Toft C, Hammarstrand L. Long-term Visual Localization using Semantically Segmented Images[J]. 2018.
[12] Schindler, A. "Vehicle self-localization with high-precision digital maps." *Intelligent Vehicles Symposium* IEEE, 2013:141-146.
[13] Zhao, Hengshuang, et al. "Pyramid Scene Parsing Network." IEEE Conference on Computer Vision and Pattern Recognition IEEE Computer Society, 2017:6230-6239.
[14] Chen, Liang Chieh, et al. "DeepLab: Semantic Image Segmentation with Deep Convolutional Nets, Atrous Convolution, and Fully Connected CRFs." IEEE Transactions on Pattern Analysis & Machine Intelligence 40.4(2016):834-848.
[15] Civera, Javier, et al. "Towards semantic SLAM using a monocular camera." *Ieee/rsj International Conference on Intelligent Robots and Systems* IEEE, 2011:1277-1284.
[16] Whelan, T, et al. "Deformation-based loop closure for large scale dense RGB-D SLAM." *Ieee/rsj International Conference on Intelligent Robots and Systems* IEEE, 2013:548-555.
[17] Rusu, Radu Bogdan, and S. Cousins. "3D is here: Point Cloud Library (PCL)." 47.10(2011):1-4.
[18] Hartley, Richard, and A. Zisserman. "Multiple view geometry in computer vision." Cambridge University Press, 2000:1865 - 1872.